\documentclass{article}
\usepackage{spconf,amsmath,graphicx}

\usepackage{enumitem}
\setlist{nosep, leftmargin=14pt}

\usepackage{mwe} 

\title{Contrastive meta-domain adaptation for robust skin lesion classification across clinical and acquisition conditions}
\name{
\parbox{\linewidth}{\centering
Rodrigo Mota \qquad Kelvin Cunha \qquad Emanoel dos Santos \qquad Fábio Papais   \qquad Francisco Filho \qquad Thales Bezerra \qquad Erico Medeiros \qquad Paulo Borba \qquad Tsang Ing Ren
}}

\address{Centro de Informática, Universidade Federal de Pernambuco, Brazil}

\begin{document}
\ninept
\maketitle
\begin{abstract}
Deep learning models for dermatological image analysis remain sensitive to acquisition variability and domain-specific visual characteristics, leading to performance degradation when deployed in clinical settings. We investigate how visual artifacts and domain shifts affect deep learning–based skin lesion classification. We propose an adaptation strategy, grounded in the idea of visual meta-domains, that transfers visual representations from larger dermoscopic datasets into clinical image domains, thereby improving generalization robustness. Experiments across multiple dermatology datasets show consistent gains in classification performance and reduced gaps between dermoscopic and clinical images. These results emphasize the importance of domain-aware training for deployable systems.
\end{abstract}

\begin{keywords}
Dermatology, Skin lesion classification, Domain shift, Continual learning
\end{keywords}
\section{Introduction}
\label{sec:intro}

Deep learning has shown strong potential for supporting skin cancer diagnosis~\cite{chan2020machine}, yet current models often fail to generalize in practical clinical settings~\cite{adamson2018machine}. This gap remains largely from biases in image acquisition and dataset composition, which shape the visual features learned during training and can lead to unreliable predictions. Although clinical workflows introduce artifacts that may aid human interpretation, they frequently degrade model performance~\cite{santos2025analysis}. Moreover, many works evaluate models using images from a single acquisition source, overlooking the variability across visual domains. As clinical deployment must account for these shifts, we assess their impact and propose strategies to mitigate knowledge forgetting.

We investigate how acquisition artifacts and domain shifts affect robustness in skin lesion classification and introduce a pipeline to address these issues: (i) a multi-transform contrastive pre-training step that aligns dermoscopic and clinical representations, reducing inter-lesion similarity and improving feature separability; and (ii) a meta-domain adaptation stage that calibrates models across datasets to enhance performance under distribution shifts while minimizing knowledge loss. Experiments across multiple dermatology datasets show reduced generalization gaps and improved reliability under clinical evaluation.

\begin{figure}
	\centering
	\includegraphics[width=1 \linewidth]{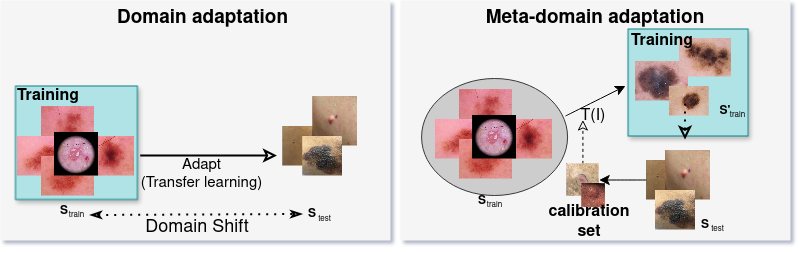}
	\caption{Overview of the domain adaptation strategy. Instead of directly transferring features from a previously trained dataset (left), we introduce meta-domain adaptation that aligns the visual representations of the source domain with those of the target domain (right).}
	\label{fig:adaptation}
\end{figure}

\section{Problem formulation}
\label{sec:format}

In supervised learning, we aim to approximate a function $f(x)$ that predicts a label $\hat{y}$ from a labeled set $S = \{(x_i, y_i)\}_{i=1}^{N}$. Ideally, $f(x)$ should emulate the human decision, $H(x) \rightarrow y$. However, since the true data distribution $p(x, y)$ is unknown, the sample set $S$ may contain biases or noise, which limits how closely $f$ can approximate $H$. We express this as $f = f' + \delta$, where $f'$ represents the ideal model and $\delta$ denotes the accumulated error induced by data variability. For vision-based models, such variability arises from acquisition or environment-related factors, including illumination, camera processing, and scene composition~\cite{volpi2021continual}. In medical imaging, particularly in clinical dermatology, such noises are further amplified by variations in imaging workflows and limited dataset diversity. Understanding and mitigating these components is essential to improving the robustness of dermatology-oriented systems.

\section{Methodology}
\label{sec:pagestyle}

\subsection{Motivation}

Despite advances in vision algorithms, adoption in dermatological practice remains limited due to variability in imaging conditions. These challenges arise from task-related noise inherent to clinical workflows. In this study, we analyze how these factors affect model decisions and introduce a domain-adaptation pipeline designed for clinical deployment. Our method adapts representations learned from large dermoscopic datasets to improve performance on smaller clinical datasets (Figure~\ref{fig:adaptation}).

\subsection{Adressing task artifacts}

\subsubsection{Clinically Related Artifacts}

Several studies attribute the limited performance of dermatology models to clinical artifacts~\cite{Pacheco2020-mn,Daneshjou2022-cd}. Datasets may exhibit variations in skin tone or underrepresentation of clinically relevant lesion types. Beyond these factors, we examine the impact of \textit{inter-lesion similarity}, which can impair model decision-making, particularly in low-resolution clinical datasets. To mitigate this effect, we employ a \textit{contrastive pre-training} strategy that enables the model to distinguish features of distinct lesions in the latent space. An EfficientNet backbone is used without the classification head, and the decoder mirrors it to reconstruct the input images. The model is trained unsupervisedly using a contrastive loss (Equation~\ref{eq:contrastive}). Where $\text{sim}(\cdot)$ denotes cosine similarity between embeddings $z_i$ and $z_j$, and $\tau$ is a temperature parameter. This process promotes discriminative representations by separating dissimilar lesions. The classification head is later fine-tuned on these pre-trained weights, enhancing feature separability and reducing inter-lesion misclassification.

\begin{equation}
\label{eq:contrastive}
\mathcal{L}_{\text{contrast}} = -\log \frac{\exp(\text{sim}(z_i, z_j)/\tau)}{\sum_{k=1}^{N}\exp(\text{sim}(z_i, z_k)/\tau)}    
\end{equation}

To enforce consistent attribution of images from the same sample under varying visual conditions, we employ a multi-transform strategy. For each image $x_i$, we apply a set of $N$ stochastic augmentations, producing a set of transformed samples $\{\hat{x}_{i}^{1}, \dots, \hat{x}_{i}^{N}\}$. The original and augmented images are encoded into a shared embedding space, yielding
$z_i = f(x_i)$,  $\hat{z}_{i}^{k} = f(\hat{x}_{i}^{k})$, $k = 1, \dots, N,$ where $f(\cdot)$ denotes the encoder network. We then compute pairwise cosine similarities between embeddings in the combined set $z_i \cup \{\hat{z}_{i}^{k}\}$. A positive mask is defined to mark pairs corresponding to different views of the same sample (i.e., original vs. augmented), while all other embeddings are treated as negatives. The training objective is a multi-positive InfoNCE loss, as in equation~\ref{eq:infonce}, where $\alpha=\frac{1}{N+1}$.

\begin{equation}
\label{eq:infonce}
\mathcal{L}_{\text{multi-contrast}} = - \alpha \sum_{k=0}^{N} 
\log \frac{\exp\left(\text{sim}( \hat{z}_{i}^{k}, z_i ) / \tau \right)}
{\sum_{j \neq i} \exp\left(\text{sim}( \hat{z}_{i}^{k}, z_j ) / \tau \right)},
\end{equation}

The $\mathcal{L}_{\text{multi-contrast}}$ encourages transformed views of the same lesion to remain close in the embedding space, while maximizing separation from embeddings of different samples. Consequently, the model learns invariant visual features that remain stable across clinical variability.

\subsubsection{Domain shift related artifacts}

Domain shift artifacts arise from discrepancies in data distributions across acquisition devices or capture environments. This phenomenon is widespread in visual dermatology assessments. Each device may apply unique preprocessing operations that alter pixel statistics, such as color correction, gamma adjustment, or denoising. Formally, a domain shift occurs when the marginal distribution of the source domain $p_{ds}(x)$ differs from that of the target domain $p_{dt}(x)$, causing models trained on $p_{ds}(x, y)$ to generalize poorly to $p_{dt}(x, y)$. Neural network-based models trained via backpropagation often suffer from catastrophic forgetting~\cite{volpi2021continual}, as their weights are optimized to minimize a loss function within the current distribution (Equation~\ref{eq:backprop}). As datasets expand across, a model may lose previously learned knowledge and corrupt its internal representations, leading to degraded performance on prior domains. 

\begin{equation}
\label{eq:backprop}
\hat{\theta}_{ds} = \min_{\theta_{ds}} \left\{ L_{o}(S_{ds}; \theta_{ds}) = -\frac{1}{|S_{ds}|}\sum_{k=1}^{|S_{ds}|} y_{k}\log\hat{y}_{k} \right\}
\end{equation}

Inspired by continual adaptation~\cite{volpi2021continual}, we introduce a guided adaptation strategy (i.e., guided-tuning) to reduce inter-domain discrepancies and mitigate catastrophic forgetting across related domains. Let a model $f_{ds}$ be initially trained on a source dataset $S_{\text{ds}}$ with parameters $\theta_{ds}$. We aim to adapt this model to a target domain $S_{dt}$ using the loss objective $L_{o}$. A calibration subset $S_{\text{cal}} \subseteq S_{dt}$ is extracted to estimate domain-specific hyperparameters $\gamma_{\text{cal}}$, which guide a pre-training step applying domain-relevant augmentations to $S_{\text{ds}}$. Let $S_{\text{cal}} \cup S_{\text{ds'}} = S_{\text{adapt}}$, we create $K$ meta-domains to simulate optimization steps guided toward the target domain ($\{\hat\theta^t_j = \theta^t - \alpha\nabla_{\theta}L_{o}(S_{\text{adapt}}; \theta^t)\}_{j=1}^{K}$). Instead of defining different loss functions or random transformations~\cite{volpi2021continual}, we reuse an $L_{o}$ across related domains, adapting knowledge to new samples while preserving essential features from previous ones by using the target distribution to guide our transformations. The model is then updated according to Equation~\ref{eq:guided-tuning}, incorporating adaptation terms into the optimization process.

\begin{multline}
\label{eq:guided-tuning}
L_{o}^{\text{total}} = L_{o}(S_{dt}; \theta^t) +\\ 
\beta_{1}\frac{1}{K}\sum_{j=1}^{K}L_{o}(S_{dt}; \hat{\theta}_{j}^{t}) +
\beta_{2}\frac{1}{K}\sum_{j=1}^{K}L_{o}(S_{\text{adapt}};\hat{\theta}_{j}^{t})
\end{multline}

\begin{figure}
	\centering
	\includegraphics[width= 0.9 \linewidth]{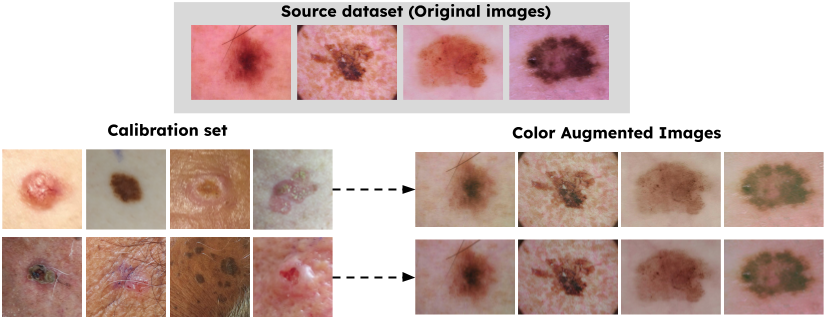}
	\caption{Color-based appearance transformations applied to the source domain to generate $K$ meta-domains, derived from randomly selected calibration subsets for domain adaptation.}
	\label{fig:transforms}
\end{figure}

We adapt the appearance by estimating global color statistics from the calibration images, transferring these properties to the source set while preserving spatial structure. Specifically, color features are obtained by estimating the mean and standard deviation in the LAB color space. These statistics are then used to parameterize color-transfer transformations implemented via Albumentations~\cite{buslaev2020albumentations}, mapping target images to the reference appearance distribution. In addition, blur characteristics are estimated using Laplacian variance and gradient-based measures, which are subsequently used to apply blur-related degradations, including gaussian, motion, and defocus blur. Standard flip and orientation transformations are also employed. All transformations are applied stochastically with a probability of 0.5 within the Albumentations pipeline. No augmentation that alters geometric or morphological properties is used, as changes in lesion structure could negatively impact diagnostic assessment. To increase appearance diversity under limited target data, the calibration set is randomly partitioned into $K=2$ meta-domains, with one selected at each training iteration. The augmented updates are weighted equally ($\beta_1 = \beta_2 = 0.5$). Figure~\ref{fig:transforms} illustrates examples of the resulting color transformations to reduce the appearance gap between the source and target domains during model adaptation.

\subsection{Dataset and models validation}

We use three datasets capturing complementary sources of image variability: \textbf{HAM10000}~\cite{tschandl2018ham10000}, which provides high-resolution dermoscopic images; \textbf{PAD-UFES-20}~\cite{Pacheco2020-mn}, consisting of clinical images acquired with smartphone cameras; and \textbf{DDI}~\cite{Daneshjou2022-cd}, a clinical smartphone dataset emphasizing diverse skin tones. These datasets enable a systematic evaluation of model generalization. Experiments are conducted using an EfficientNet backbone. Each proposed component is first evaluated independently to quantify its contribution across datasets, followed by an integrated evaluation of domain-aware processing. Performance is reported using accuracy and F1-score~\cite{tschandl2018ham10000,Pacheco2020-mn}, following the original dataset protocols. For HAM10000, we use the standard training split with $80\%$/$20\%$ train/validation and the original test set ($1512$ images). For PAD-UFES-20 and DDI, a $70\%$/$15\%$/$15\%$ split is used for training, validation, and testing.

\section{Experiments and discussions}

We trained the model on the HAM10000 ~\cite{tschandl2018ham10000}, a large collection captured with magnifying devices that highlight fine lesion details. These models were optimized to classify in dermoscopic images. However, our goal is to apply them in clinical scenarios, using datasets such as PAD-UFES-20~\cite{Pacheco2020-mn} and DDI~\cite{Daneshjou2022-cd}. These datasets are comparatively smaller, noisier, and contain multiple visual artifacts. Therefore, in our experiments, we aim to adapt the feature representations learned from HAM10000 to the clinical domain while mitigating knowledge degradation during adaptation. 

\subsection{Constrative pre-training}

\begin{figure}
	\centering
	\includegraphics[width= 0.8\linewidth]{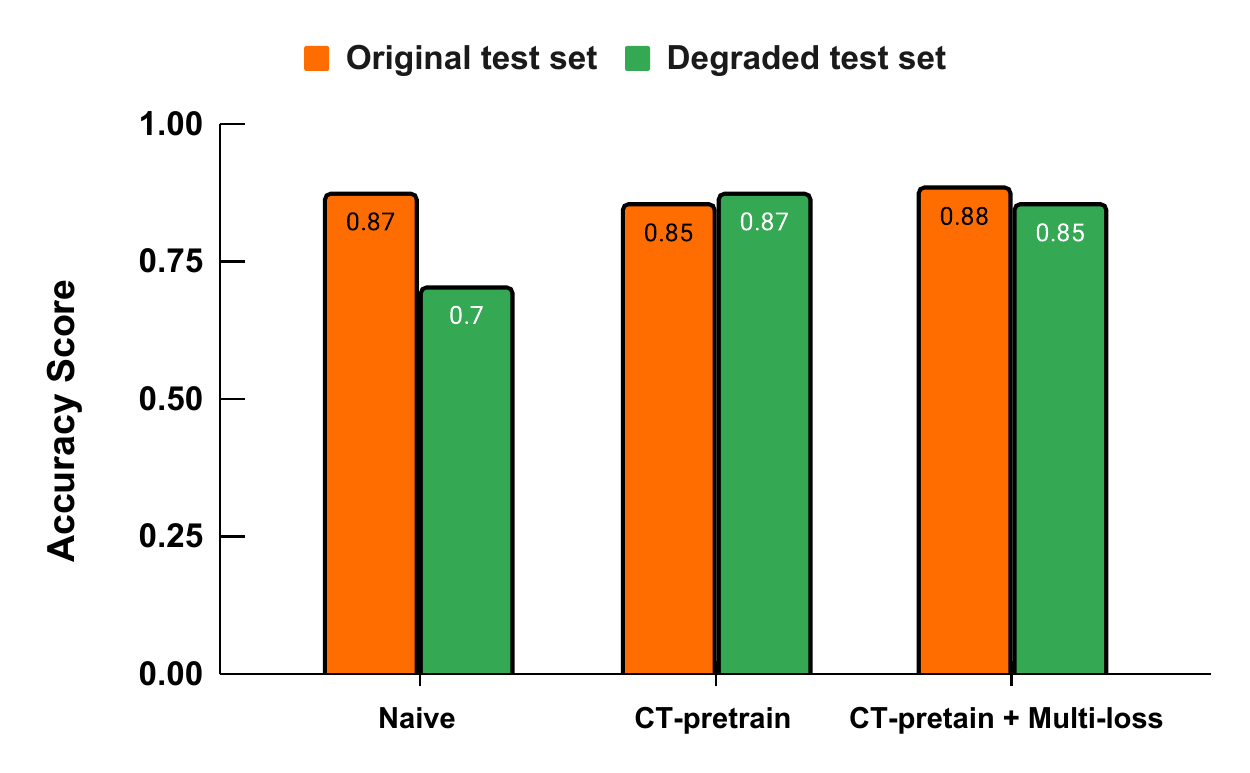}
	\caption{Contrastive pre-training on HAM10000 test set. Naive training is compared with contrastive optimization to mitigate robustness degradation in visual domain features.}
	\label{fig:ct}
\end{figure}

We evaluate contrastive pre-training (CT-pretrain) on the HAM10000 dataset by comparing it with standard supervised backpropagation (Naive). Models are trained on the HAM10000 training split and evaluated on both the original test set and a degraded version that simulates realistic clinical artifacts, including blur, sensor noise, illumination shifts, motion blur, and overexposure. A multi-transform contrastive loss is used during pre-training to improve robustness to imaging variability. While naive training exhibits substantial performance degradation under perturbations, CT-pretrain maintains more stable performance across both evaluation settings. Figure~\ref{fig:ct} summarizes the impact of degradations, highlighting the robustness gains obtained with contrastive pre-training.

\subsection{Domain adaptation to clinical datasets}

\begin{table*}[]
\centering
\caption{Comparison of naive training, traditional fine-tuning with augmentation, and the proposed contrastive pre-training (CT-pretrain) and guided-tuning (GT) strategies. We replicate the baseline methods reported in~\cite{Pacheco2020-mn,Daneshjou2022-cd} and evaluate all approaches on their respective datasets. Metrics include accuracy, precision, recall, and F1-score.}
\label{tab:results}
\resizebox{0.7\textwidth}{!}{\begin{tabular}{c|cccccccc}
 & \multicolumn{4}{c}{PAD-UFES-20~\cite{Pacheco2020-mn}} & \multicolumn{4}{c}{DDI~\cite{Daneshjou2022-cd}} \\ \hline
 & ACC & F1 & Precision & \multicolumn{1}{c|}{Recall} & ACC & F1 & Precision & Recall \\ \hline
Pacheco et. al.~\cite{Pacheco2020-mn}(only image) & 0.69 & 0.7 & 0.73 & \multicolumn{1}{c|}{0.69} & - & - & - & - \\
Daneshjou et al.~\cite{Daneshjou2022-cd} & - & - & - & \multicolumn{1}{c|}{-} & 0.77 & 0.76 & 0.76 & 0.77 \\ \hline
Naive (backprop) & 0.35 & 0.38 & 0.38 & \multicolumn{1}{c|}{0.39} & 0.12 & 0.05 & 0.04 & 0.11 \\
Naive + Fine-tuning (FT) & 0.71 & 0.64 & 0.6 & \multicolumn{1}{c|}{0.69} & 0.5 & 0.53 & 0.54 & 0.53 \\
Naive + CT-pretrain & 0.76 & 0.72 & 0.71 & \multicolumn{1}{c|}{0.73} & 0.49 & 0.3 & 0.28 & 0.33 \\
\begin{tabular}[c]{@{}c@{}}Naive + FT \\ + random augment\end{tabular} & 0.72 & 0.71 & 0.71 & \multicolumn{1}{c|}{0.72} & 0.58 & 0.5 & 0.54 & 0.48 \\ \hline
\textbf{Guided-tuning (GT)} & 0.83 & 0.82 & 0.81 & \multicolumn{1}{c|}{0.83} & 0.79 & 0.79 & \textbf{0.81} & 0.79 \\
\textbf{CT-pretrain + GT} & \textbf{0.88} & \textbf{0.84} & \textbf{0.83} & \multicolumn{1}{c|}{\textbf{0.85}} & \textbf{0.79} & \textbf{0.81} & 0.8 & \textbf{0.81}
\end{tabular}}
\end{table*}

\begin{figure}
	\centering
	\includegraphics[width=0.8\linewidth]{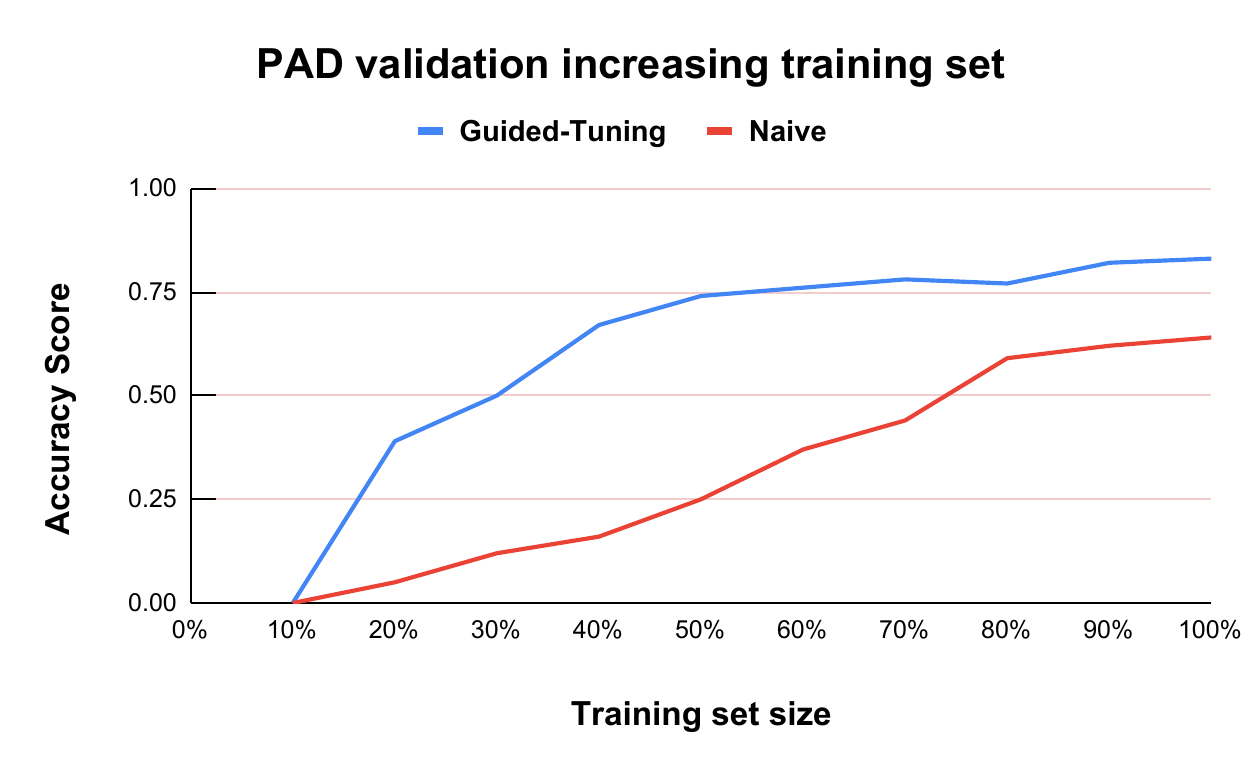}
	\caption{Validation of guided-tuning on PAD increasing the training set size.}
	\label{fig:abl}
\end{figure}

We next evaluate guided tuning (GT) under distribution shifts and its complementarity with contrastive pre-training (CT). Table~\ref{tab:results} reports the effects of GT and CT relative to the original CNN baselines~\cite{Pacheco2020-mn, Daneshjou2022-cd} and our architecture trained with naive backpropagation. Using the proposed architecture, we compare standard fine-tuning and augmentation with CT and GT. While fine-tuning and augmentation improve performance over naive training, CT yields larger gains in clinical settings characterized by noise, artifacts, and acquisition variability, which is more prevalent than in dermoscopic imaging. GT further improves robustness by leveraging dermoscopic priors to guide feature adaptation in the clinical domain, promoting representation alignment while preserving lesion-specific cues. Notably, GT enables stable adaptation with limited target-domain data (Figure~\ref{fig:abl}).

\begin{figure}
	\centering
	\includegraphics[width=0.8\linewidth]{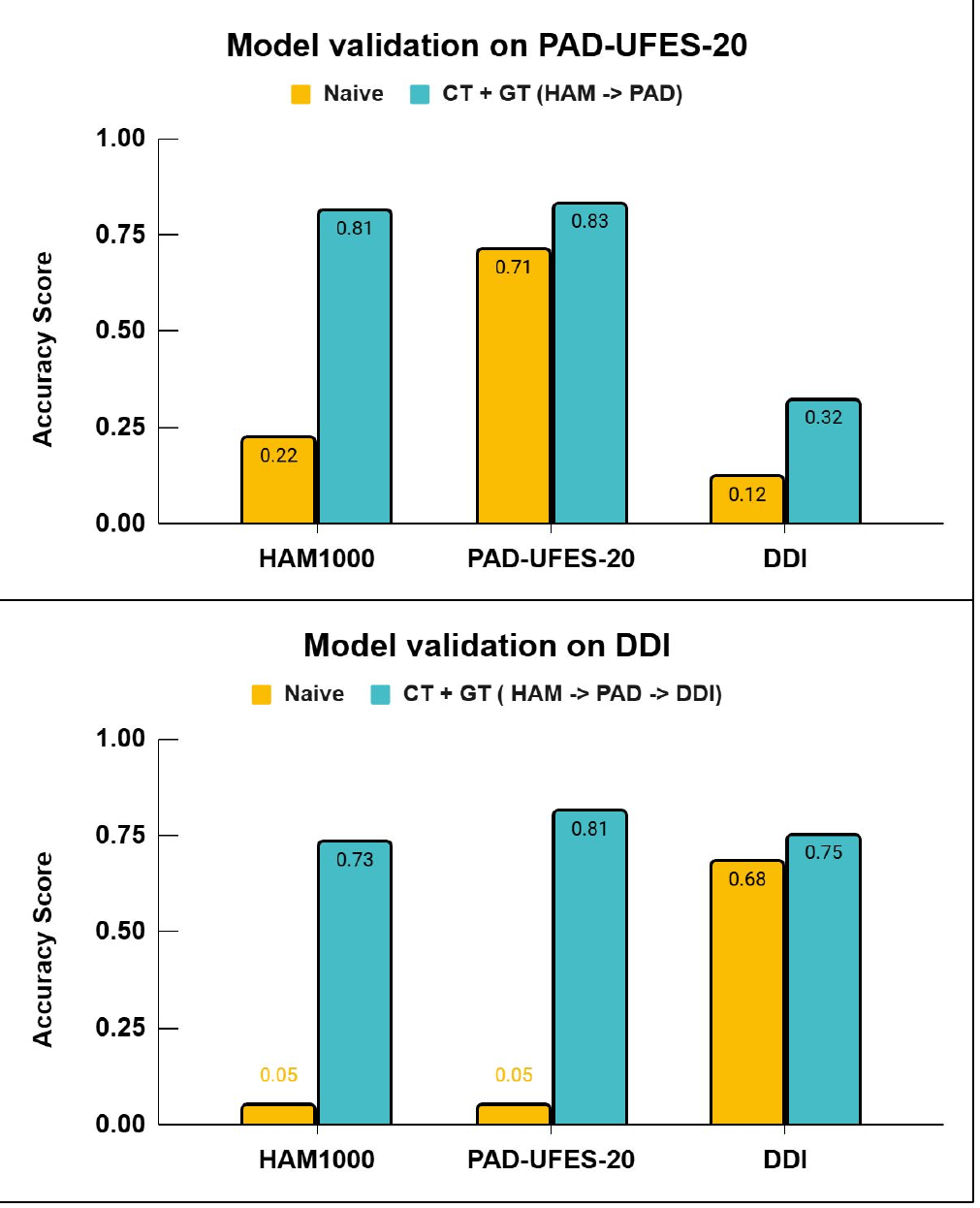}
	\caption{Comparison between the guided-tuning and naive training. Naive backpropagation adapts to the current domain but exhibits forgetting of previously learned domains and poor generalization to unseen ones.}
	\label{fig:val-pad-ddi}
\end{figure}

Figure~\ref{fig:val-pad-ddi} further demonstrates the GT's benefits. For both PAD and DDI-trained models, GT and naive training yield similar performance on their respective training domains. However, only GT preserves performance on previously learned domains (e.g., HAM10000 for models adapted to PAD, and PAD for models adapted to DDI). Thus, even as additional data are incorporated, GT prevents catastrophic forgetting and maintains useful knowledge in different visual domains. This contributes directly to building reliable AI systems for clinical skin-lesion assessment.
 
\section{Conclusion}

In this work, we introduce a training strategy to mitigate performance degradation in clinical models for dermatological lesion assessment. Traditional machine learning approaches often struggle with domain variations, which critically impact real-world reliability. To address this, we propose a contrastive pre-training strategy that enhances robustness to image degradation and acquisition noise by improving feature disentanglement across similar lesions. Furthermore, we present a guided training optimization scheme for continual learning, enabling models to adapt to new domains while preserving knowledge from previously trained ones. By leveraging high-quality features from related domains, our approach helps build more reliable and deployable clinical models for skin lesion analysis. 

\section{Compliance with ethical standards}
\label{sec:ethics}

This research study was conducted retrospectively using human subject data made available in open access~\cite{tschandl2018ham10000,Pacheco2020-mn,Daneshjou2022-cd}. Ethical approval was not required, as confirmed by the license attached to the open-access data.

\section{Acknowledgments}
\label{sec:acknowledgments}

This work was partially supported by INES.IA (National Institute of Science and Technology for Software Engineering Based on and for Artificial Intelligence) www.ines.org.br, CNPq grant 408817/2024-0. The project was supported by the Ministry of Science, Technology, and Innovation of Brazil, with resources from Law No. 8,248, dated October 23, 1991, under the scope of the PPI-SOFTEX, coordinated by Softex and published under RESIDÊNCIA EM TIC 63 – ROBÓTICA E IA – FASE II, DOU 23076.043130/2025-27. 

\bibliographystyle{IEEEbib}
\bibliography{references}

\end{document}